\documentclass[10pt,twocolumn,letterpaper]{article}
\usepackage[pagenumbers]{cvpr} 
\usepackage{graphicx}
\usepackage{amsmath}
\usepackage{amssymb}
\usepackage{booktabs}
\usepackage{enumitem}
\setlist{leftmargin=5.5mm}

\usepackage{amsmath}
\usepackage{bm}
\usepackage[utf8]{inputenc} 
\usepackage[T1]{fontenc}    
\usepackage{hyperref}       
\usepackage{url}            
\usepackage{nicefrac}       
\usepackage{microtype}      
\usepackage{xcolor}         

\def\eqref#1{equation~\ref{#1}}

\def\1{\bm{1}}

\def\vx{{\bm{x}}}

\def\vz{{\bm{z}}}

\def\mG{{\bm{G}}}

\def\mX{{\bm{X}}}

\def\mZ{{\bm{Z}}}

\DeclareMathAlphabet{\mathsfit}{\encodingdefault}{\sfdefault}{m}{sl}
\SetMathAlphabet{\mathsfit}{bold}{\encodingdefault}{\sfdefault}{bx}{n}

\newcommand{\Cov}{\mathrm{Cov}}

\DeclareMathOperator{\CosSim}{CoSim}

\usepackage{tikz}

\usepackage{hyperref}
\usepackage{url}
\usepackage{multirow}
\usepackage{enumitem}

\usepackage{amsthm}
\usepackage{amssymb}
\usepackage{mathtools}
\theoremstyle{plain}

\usepackage{varwidth}
\usepackage{enumitem}
\usepackage{tikz}
\usetikzlibrary{positioning}
\usetikzlibrary{shapes}
\usetikzlibrary{fit}
\usetikzlibrary{calc}

\usepackage{listings}
\usepackage{xcolor}
\usepackage{courier}

\definecolor{codegreen}{rgb}{0,0.6,0}
\definecolor{codegray}{rgb}{0.5,0.5,0.5}
\definecolor{codeblack}{rgb}{0.,0.,0.}
\definecolor{codepurple}{rgb}{0.58,0,0.82}
\definecolor{backcolour}{rgb}{0.95,0.95,0.92}

\lstdefinestyle{mystyle}{
    backgroundcolor=\color{backcolour},   
    commentstyle=\color{codegreen},
    keywordstyle=\color{codeblack},
    numberstyle=\tiny\color{codegray},
    stringstyle=\color{codepurple},
    basicstyle=\ttfamily\footnotesize,
    breakatwhitespace=false,         
    breaklines=true,                 
    captionpos=b,                    
    keepspaces=true,                 
    numbers=left,                    
    numbersep=5pt,                  
    showspaces=false,                
    showstringspaces=false,
    showtabs=false,                  
    tabsize=2,
    aboveskip=0pt,
    belowskip=-3pt
}

\usepackage[capitalize]{cleveref}
\crefname{section}{Sec.}{Secs.}
\Crefname{section}{Section}{Sections}
\Crefname{table}{Table}{Tables}
\crefname{table}{Tab.}{Tabs.}

\usepackage{float}
\begin{document}

\title{Towards Democratizing Joint-Embedding Self-Supervised Learning}

\author{Florian Bordes$^{1,2}$, Randall Balestriero$^{2}$, Pascal Vincent$^{1,2}$ \\
  $^1$Mila, Université de Montréal, $^2$Meta AI}
  
\maketitle

\vspace{-0.1cm}

\begin{abstract}
Joint Embedding Self-Supervised Learning (JE-SSL) has seen rapid developments in recent years, due to its promise to effectively leverage large unlabeled data. The development of JE-SSL methods was driven primarily by the search for ever increasing downstream classification accuracies, using huge computational resources, and typically built upon insights and intuitions inherited from a close parent JE-SSL method. This has led unwittingly to numerous pre-conceived ideas that carried over across methods e.g. that SimCLR requires very large mini batches to yield competitive accuracies; that strong and computationally slow data augmentations are required.
In this work, we debunk several such ill-formed a priori ideas in the hope to unleash the full potential of JE-SSL free of unnecessary  limitations.
In fact, when carefully evaluating performances across different downstream tasks and properly optimizing hyper-parameters of the methods, we most often --if not always-- see that these widespread misconceptions do not hold. For example we show that it is possible to train SimCLR to learn useful representations, while using a single image patch as negative example, and simple Gaussian noise as the only data augmentation for the positive pair. Along these lines, in the hope to democratize JE-SSL and to allow researchers to easily make more extensive evaluations of their methods, we introduce an optimized PyTorch library for SSL \url{https://github.com/facebookresearch/FFCV-SSL}.
\end{abstract}

\begin{figure}
    \centering
    \includegraphics[width=\linewidth]{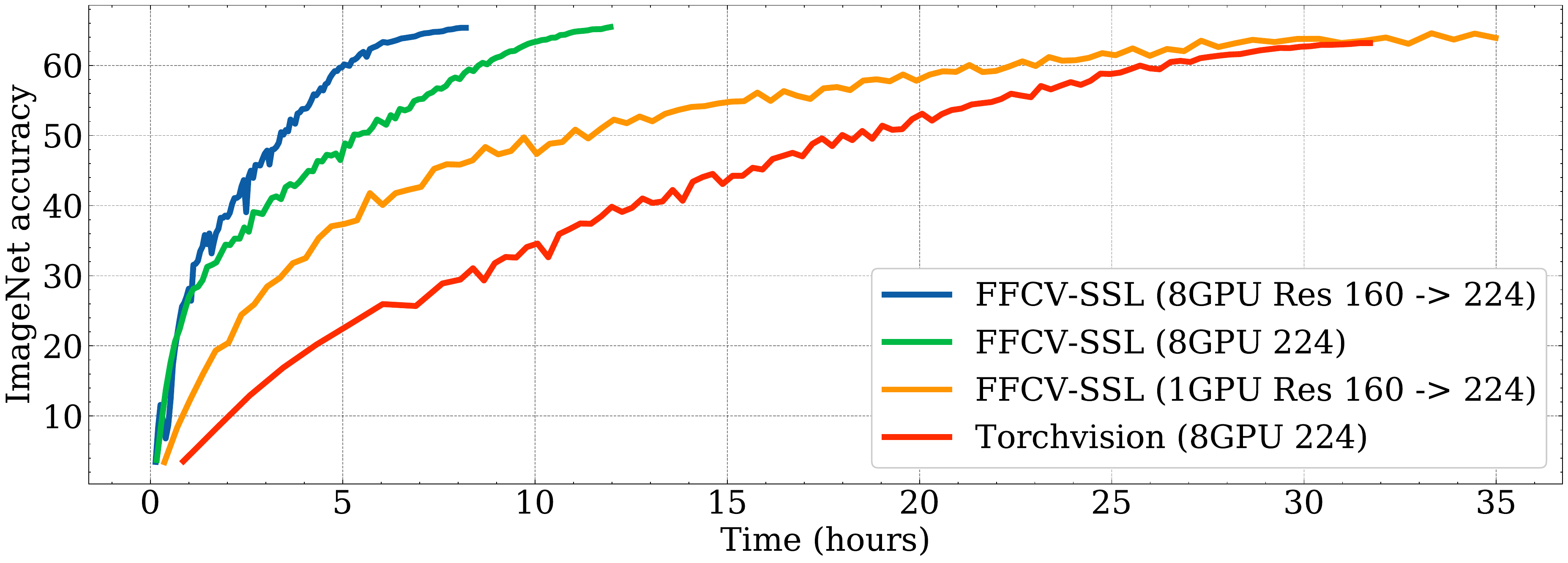}
    \caption{ImageNet validation accuracy (y-axis) during training of SimCLR with respect to the training time (x-axis). FFCV-SSL is our proposed library that is specifically optimized for Self-Supervised Learning, and that extends the original FFCV library \cite{leclerc2022ffcv}. We compare FFCV-SSL with torchvision using various image's resolution (224 means that a fixed resolution size of 224x224 is used when cropping the images while 160 -> 224 means that the resolution is increasing during training from 160x160 to 224x224) --no other changes have been applied in the implementation and the same hardware (GPU A100) is employed. Enabled by FFCV-SSL, we are able to perform thorough empirical investigations against preconceived failure modes of SSL models for which we obtain novel conclusions e.g. \textbf{1) SimCLR can perform equally well with small or large mini-batch training, 2) strong data-augmentations are not always necessary and cropping+grayscale is enough to reach competitive performances across SSL methods, and 3) it is now possible to train SSL method, e.g. SimCLR in this figure, using only 1 GPU in a reasonable amount of time.}}
    \label{fig:ffcv_vs_torchvision}
\end{figure}

\vspace{-0.3cm}
\section{Introduction}
\label{sec:intro}
\vspace{-0.1cm}

Interest in Self-Supervised Learning (SSL) has increased steadily since the work of \cite{chen2020simclr}. By using very specific sets of data augmentations to design positives pairs of examples, as well as using large mini-batch of images to define the negatives examples, \cite{chen2020simclr} demonstrated the competitiveness of SSL with respect to  supervised baselines. Since then, other works tried to build upon the contrastive method of \cite{chen2020simclr} by either increasing the scale \cite{chen2020simclrv2}, improving the negative example sampling scheme using a buffer \cite{chen2020simclrv2} or by using new data augmentations \cite{Dwibedi2021WithAL}. Despite their successes, contrastive methods are subject to several frequent criticisms. The most common one is the assumed need for a large number of negative examples, i.e. large batches, which limits the accessibility of contrastive method training to researchers who have access to massive and costly hardware resources. Another common criticism is the requirement for a very specific set of hand-crafted data augmentation to make such methods work. Moreover, in many instances, the use of these data augmentations can considerably increase the training time.

The computational burden of SSL directly impairs its widespread adoption since access to multi-GPU and large scale training is not guaranteed, and because most available resources are turned towards producing yet better performing SSL variants. Two direct consequences are that (i) there does not exist practical guideline in the literature prescribing how to perform more computationally friendly SSL even if it is at the cost of slightly reduced top-1 performances, and (ii) that successively refined SSL methods rarely spend resources to question or contest the empirical guidelines that were developed in previous studies. Those two points also interplay with each other since (ii) for example commonly prescribes the use of color-jitter data-augmentation with the goal to produce state-of-the-art performances, but this augmentation is also among the most computationally expansive to apply in practice.

In this paper, we first show that several such widely held ideas concerning Joint Embedding Self-Supervised Learning methods (JE-SSL) are misleading, and are an obstacle for the democratization of JE-SSL methods. 
Here are further illustrations of popular misconceptions that will be debunked in this paper:
\begin{itemize}
    \item \textit{"We find that, when the number of training epochs is small (e.g. 100 epochs), larger batch sizes have a significant advantage over the smaller ones."} \cite{chen2020simclr} \textit{"SimCLR and SwAV both require a large batch (e.g., 4096) to work well."} \cite{chen2020simsiam}
    \textit{"Contrastive methods suffer from the need of a lot of negative examples which can translate into the need for very large batch sizes"}\cite{bardes2016vicreg}
    \item \textit{We find that [Byol] is not robust to removing some types of data augmentations, like SIMCLR} \cite{grill2020byol}
\end{itemize}
By reproducing many experiments of the original work of \cite{chen2020simclrv2} we will be able to debunk most of the aforementioned a prioris.
In fact, it appears that some of the most influential papers in the field may have paid insufficient attention to a fundamental aspect in empirical machine learning research: hyper-parameters tuning and diversity in the evaluation protocol. Our main takeways will be that JE-SSL need special care as their performance vary greatly with respect to the employed loss' hyper-parameters, and that solely looking at Imagenet-1k\cite{deng2009imagenet} downstream performance is often misleading when it comes to measuring the quality of a learned representation.
Being free of such misconceptions, we then explore a more extreme scenario. We train SimCLR with a single negative example which is taken from a small patch of the positive pair. Doing this with only a Gaussian noise as data augmentation for the positive pair, leads on several downstream tasks to results that are very close to the SimCLR baseline.

All of our empirical analysis is enable by FFCV-SSL which we developed specifically to reduce data loading overhead when training JE-SSL methods, and is based on the fast data loading library FFCV \cite{leclerc2022ffcv}.
By using FFCV-SSL\url{https://github.com/facebookresearch/FFCV-SSL}, we were able to run SSL experiment 3 times faster than before.

\vspace{-0.2cm}
\section{Joint-Embedding Self-Supervised Methods and Notations}
\vspace{-0.1cm}

JE-SSL relies on processing multiple --semantically related-- views of a same input through a nonlinear mapping, commonly a deep network (DN), and enforcing that the produced representations of those views are close to each other. This matching is enforced through the {\em positive term} of the loss function, all while preventing the DN's mapping to collapse e.g. to a constant function through a {\em collapse prevention term}. Different flavors of positive and collapse prevention terms lead to different JE-SSL methods.

{\bf Dataset, Data-Augmentation and Relation Matrix Notations.}~Regardless of the loss and method employed, SSL relies on having access to a set of observations i.e. input samples $\mX\triangleq [\vx_1, \dots, \vx_{N}]^T\in\mathbb{R}^{N \times D}$ and a known positive relationship between those samples e.g. in the form of a {\em symmetric} matrix $\mG \in (\mathbb{R}^+)^{N \times N}$ where $(\mG)_{i,j}>0$ iff samples $\vx_i$ and $\vx_j$ are semantically related, and with $0$ in the diagonal. Commonly, one is only given a dataset $\mX'$ of size $N'$, and artificially constructs the JE-SSL dataset $\mX$ and $\mG$ from augmentations of $\mX'$, e.g. rotated (for images) or noised versions of the original samples, obtained through transformations $t$ drawn from some distribution $\mathcal{T}$. Of course, such transformations are designed to preserve the semantic information of the original inputs as those are employed to determine the positive views of the inputs as in
\begin{gather*}
    \begin{bmatrix}
    \vx_1^T\\
    \vdots\\
    \vx_{N'}^T\\
    \end{bmatrix} \xrightarrow{t_{i,j} \sim \mathcal{T}, \forall i,j}
 \begin{bmatrix}
    t_{1,1}(\vx_1)^T\\
    t_{1,2}(\vx_1)^T\\
    \vdots\\
    t_{N',1}(\vx_{N'})^T\\
    t_{N',2}(\vx_{N'})^T\\
    \end{bmatrix}
    ,
\end{gather*}
with $(\mG)_{i,j}=1_{\{j-1=i\}} + 1_{\{j+1=i\}}$ and where in this case each original input is used to general two positive views.
Lastly, $\mZ\in\mathbb{R}^{N \times K}$ denotes the matrix of feature maps obtained from a model $f_{\theta}: \mathbb{R}^{D} \mapsto \mathbb{R}^{K}$ ---commonly a Deep Network--- as $\mZ\triangleq [f_{\theta}(\vx_1), \dots, f_{\theta}(\vx_{N})]^T$.

{\bf VICReg}'s loss \cite{bardes2016vicreg} is defined as a function of $\mX$ and $\mG$ in the following triplet loss
\begin{gather}
\mathcal{L}=\alpha \sum_{k=1}^{K}{\rm relu}\left(1-\sqrt{\Cov(\mZ)_{k,k}}\right)+ \beta \sum_{j\not = k}\Cov(\mZ)^2_{k,j}\nonumber\\+\frac{\gamma}{N} \sum_{i=1}^{N}\sum_{j=1}^{N}(\mG)_{i,j}\|\mZ_{i,.}-\mZ_{j,.}\|_2^2.\label{eq:VICReg}
\end{gather}
We will refer to each term in \cref{eq:VICReg} as $\mathcal{L}_{\rm var}$, $\mathcal{L}_{\rm cov}$, and $\mathcal{L}_{\rm inv}$ respectively.

{\bf SimCLR}'s loss \cite{chen2020simclr} is slightly different and first produces an estimated relation matrix $\widehat{\mG}(\mZ)$ \cite{balestriero2022contrastive} that is compared to the ground-truth relation matrix $\mG$ via
\begin{align}
    (\widehat{\mG}(\mZ))_{i,j}&=\frac{e^{\CosSim(\vz_i,\vz_j)/\tau}}{\sum_{j=1,j\not = i}^{N}e^{\CosSim(\vz_i,\vz_j)/\tau}},\nonumber\\
    \mathcal{L}&=-\sum_{i=1}^{N}\sum_{h=1}^{N}(\mG)_{i,j}\log(\widehat{\mG}(\mZ))_{i,j}.\label{eq:simclr}
\end{align}
where CoSim denotes the cosine similarity, and $\tau>0$ is a temperature parameter. The only difference between SimCLR and variants s.a. NNCLR \cite{Dwibedi2021WithAL} lies in how one defines $\mG$. Hence, although we will particularly focus on SimCLR, our findings should easily extend to such variants.

{\bf BarlowTwins}'s loss \cite{zbontar2021barlow} proposes yet a slightly different approach where $\vz_i$ must be close to $\vz_j$ if $\mG_{i,j}>0$. They do so with different flavors of losses and constraints to facilitate training.
Hence, and for these models only, it is common to explicitly group $\mX$ into two subsets $\mX_{\rm left}$ and $\mX_{\rm right}$ based on $\mG$  so that $((\mX_{\rm left})_n,(\mX_{\rm right})_n),\forall n$ are all the positive pairs from $(\mX,\mG)$ \color{black}. This does not lose any generality. In fact, suppose that we have $5$ samples $a,b,c,d,e$, and that $\mG$ says that $a,b,c$ are related to each other, and that $d,e$ are related to each other. Then, we can create the two data matrices as
\begin{align*}
 \mX_{\rm left}=[a,a,b,b,c,c,d,e],\;\mX_{\rm right}=[b,c,a,c,a,b,e,d].
\end{align*} 
Once the two (left/right) views are obtained, the corresponding embeddings $\mZ_{\rm left},\mZ_{\rm right}$ can be computed and the BarlowTwins is then defined as
\begin{gather}
    \mathcal{L}=\sum_{k=1}^{K}(\CosSim((\mZ_{\rm left})_{.,k},(\mZ_{\rm right})_{.,k})-1)^2\nonumber\\+\alpha  \sum_{k=1}^{K}\sum_{k'=1,\not= k}^{K}\CosSim((\mZ_{\rm left})_{.,k},(\mZ_{\rm right})_{.,k'})^2.\label{eq:BT}
\end{gather}
where $\CosSim$ computes the cosine similarity between the two input vectors. One should notice that those terms correspond to the cross-correlation matrix between the two embeddings.\color{black}

{\bf Projector Networks} are multilayer perceptrons (MLPs) that are added on top of the DN "backbone" model that one aims to train with JE-SSL. Post-training, the projector network is removed; giving back the original DN's architecture but with much improved performance compared to not employing a projector network. This technique popularized by \cite{chen2020simclr} introduces additional hyper-parameters to tune e.g. the depth and width of that MLP. It also makes it less clear what in truth is learned by the DN backbone (as this is not the representation level on which the training objective is applied). \cite{RCDM,guillotine} demonstrated that one benefit of using a projector is that it serves as a buffer to absorb the bias of possibly misspecified data-augmentations and/or suboptimal JE-SSL loss hyper-parameters.

\vspace{-0.2cm}
\section{Debunking Popular Myths about Joint-Embedding Failure Cases}
\label{sec:debunking}

Due to the large number of hyper-parameters that JE-SSL rely on and the high computation cost required to train these models with existing software, most novel methods only explore a small part of the whole hyper-parameter space. For example only comparing with previously reported results using the same set-ups. For example, \cite{chen2020simclr, zbontar2021barlow} study the impact of the batch size but keep all other hyper-parameters fixed. Although such sensitivity experiments are useful in many ways, they also risk leading to misconceptions on the failure cases of JE-SSL. This is what we propose to investigate in this section. Surprisingly, we will be able to debunk empirically several observations that were put forward in multiple previous studies, ultimately showcasing that most JE-SSL methods are actually much more similar to one another than previously thought, and suffer much less dramatic failures than previously reported.

\subsection{The Impact of Mini-Batch Size for SimCLR and BarlowTwins}

Recall from \cref{eq:simclr} that SimCLR uses negative examples in the denominator term of its loss. It has been reasoned that the role of this term is to perform negative sampling \cite{khosla2020supervised} and thus that is will only be effective when the number of samples i.e. mini-batch size is large. Equivalently, BarlowTwins (recall \cref{eq:BT}) estimates correlation along the sample dimensions and thus also is expected to benefit from large mini-batch size for more accurate estimation. In this section, we propose to refute that large mini-batch sizes are required to successfully train JE-SSL with these two methods.

The belief that many methods rely on large mini-batch size has been mentioned in many recent studies, e.g. \cite{chen2020simsiam}. This belief was confirmed by one of the Figure in \cite{chen2020simclr} that display a 7\% accuracy gap on ImageNet between a model trained with a small (256) versus large batch size (8192). However, an important point that people often miss when reading \cite{chen2020simclr} is the critical influence of the optimizer and the choice of the learning rate when training with different batch size. In their appendix, \cite{chen2020simclr} show that when using a better optimization technique, the gap in performances between bigger and smaller batch size are significantly reduced. 

\begin{figure}
    \centering
    \includegraphics[scale=0.4]{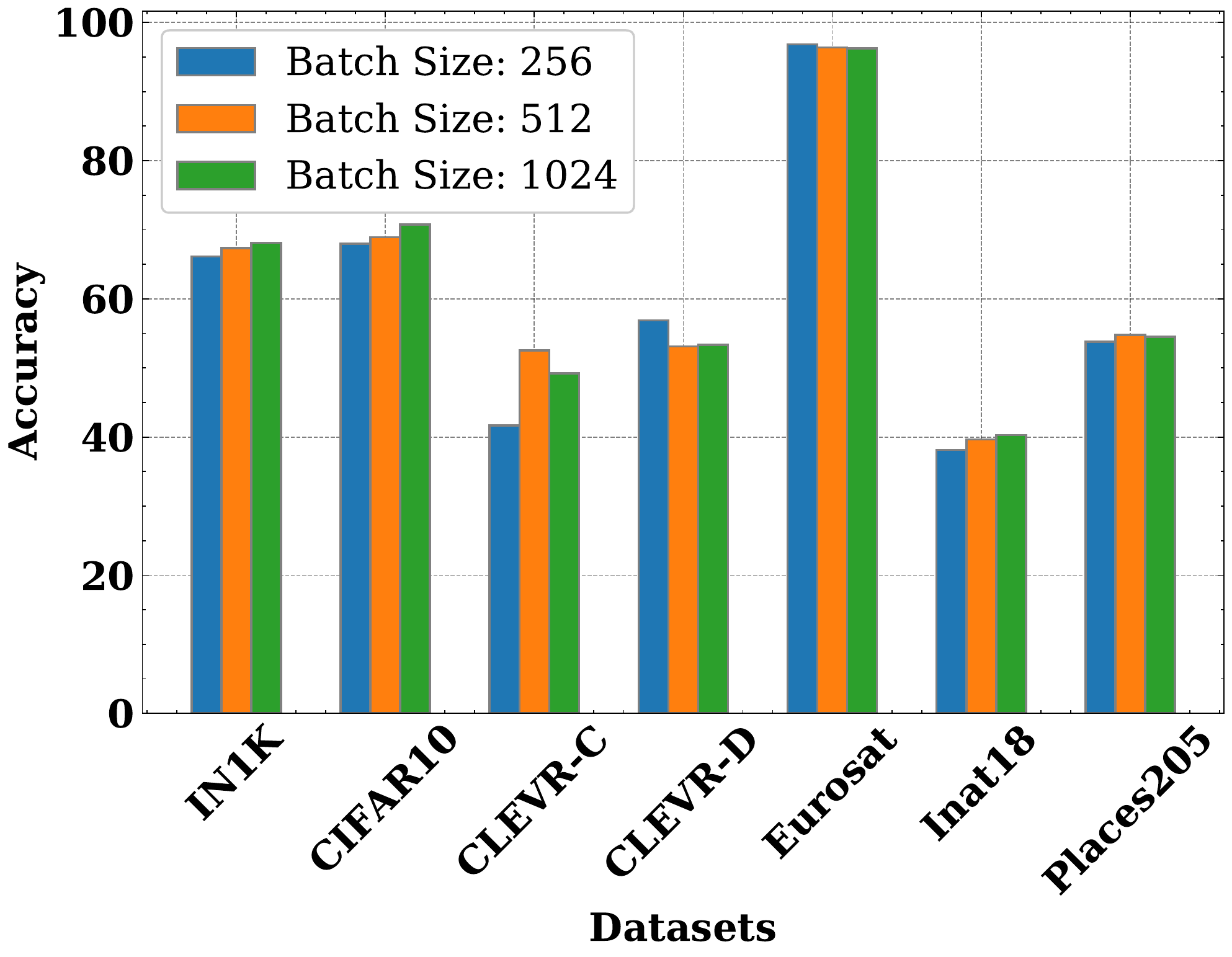}
    \caption{Accuracy across different downstream tasks given by probing SimCLR representations trained with different batch size. CLEVR-C corresponds to the task of counting the number of objects in the images whereas CLEVR-D corresponds to the task of estimating the distance between objects,. Even if the performances on ImageNet(IN1K) are better with a larger batch size, it's not necessarily the case for every downstream tasks.}
    \label{fig:datasets_vs_batchsize_simclr}
\end{figure}

\paragraph{The impact of the downstream task} One caveat of the batch size analysis in \cite{chen2020simclr, zbontar2021barlow} is that it concerns only the performances on ImageNet. Since one of the main motivation behind SSL is to learn a model whose representations can generalize to different tasks, we analyse the performances with different batch sizes across several downstream tasks: ImageNet-1K\cite{imagenet}, CIFAR10\cite{cifar10}, CLEVR \cite{johnson2017clevr}, Eurosat \cite{Eurosat}, Inaturalist \cite{Inat} and Places\cite{place205}. In Figure \ref{fig:datasets_vs_batchsize_simclr}, we plot the performances with SimCLR trained with several batch sizes. We observe that there is a small gain when using larger batch size with SimCLR on ImageNet (IN1K) however the benefit of using larger batch size is not consistent across all of the downstream tasks. The performances on the Eurosat, Places and CLEVR dataset are not better with a large batch size. 

This suggests that researchers should exert caution before broadly declaring that a given method requires a large batch size. While the claim may be empirically verified in one setting (ImageNet), it might not be true in other scenarios.

\begin{figure}
    \centering
    \includegraphics[scale=0.4]{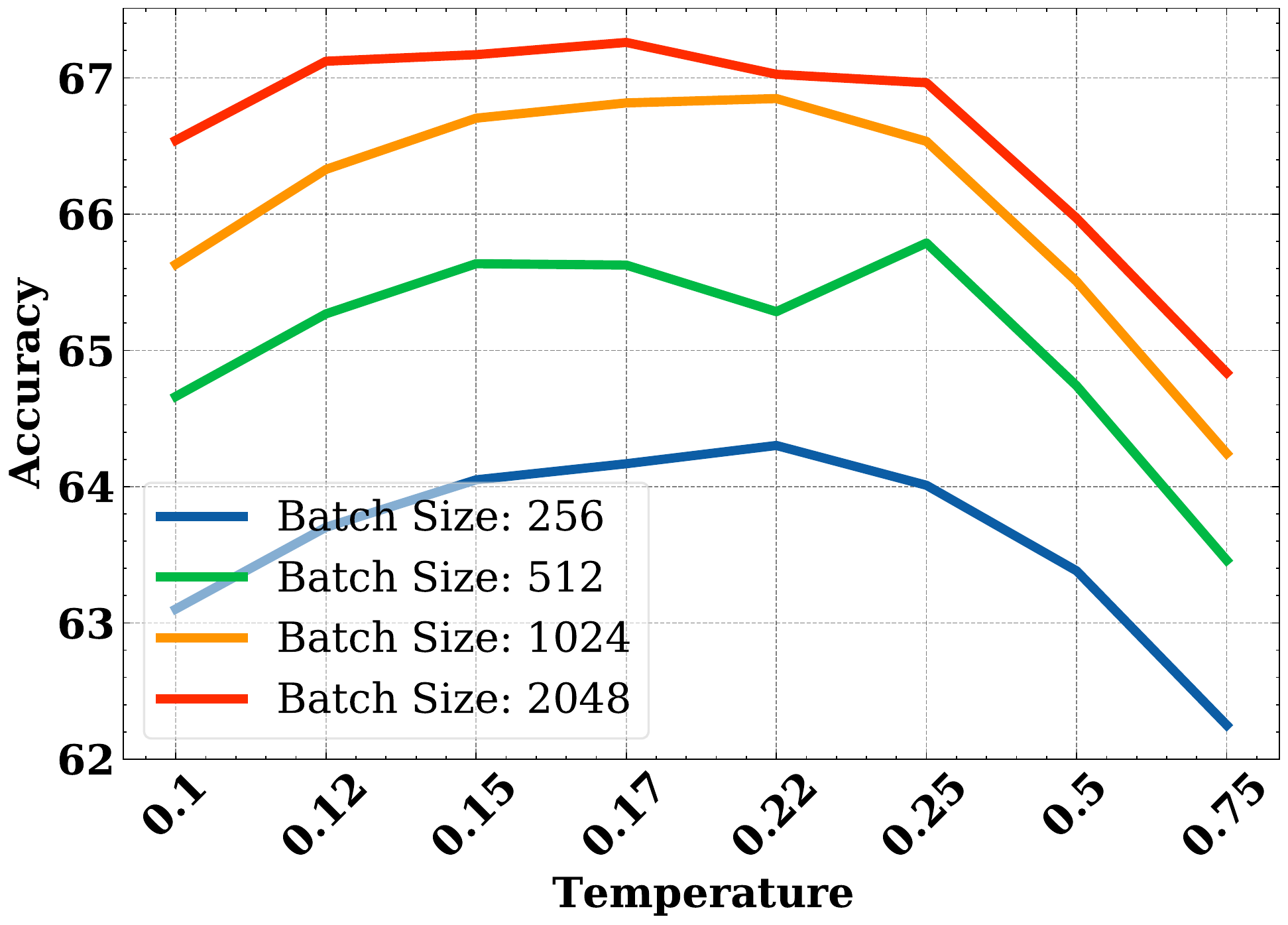}
    \caption{Validation accuracy on ImageNet with respect to the temperature parameters in SimCLR and the batch size. When having a temperature of 0.1, the gap between a large batch size can be as high as 4\%. However, when carefully running a grid search, we observe that the optimal temperature might not be the same depending of the batch size.}
    \label{fig:simclr_temperature}
\end{figure}

\paragraph{The impact of hyper-parameters of the loss } Another important aspect that interacts with the batch size is the optimization of the hyper-parameters of the SSL loss. Because of the supposed large computational requirements of SSL, most researchers have performed their experiments by varying only a given factor at a time without cross-validation. For example, \cite{chen2020simclr, zbontar2021barlow} study the influence of the batch size only for a given temperature for SimCLR and a given hyper-parameter lambda for Barlow Twins. However the influence of the temperature in \cref{eq:simclr} also depends on the batch size which has a direct impact on the scale.  \cref{fig:simclr_temperature} shows the impact of the batch size with respect to SimCLR temperature.

\begin{figure}
    \centering
    \includegraphics[scale=0.4]{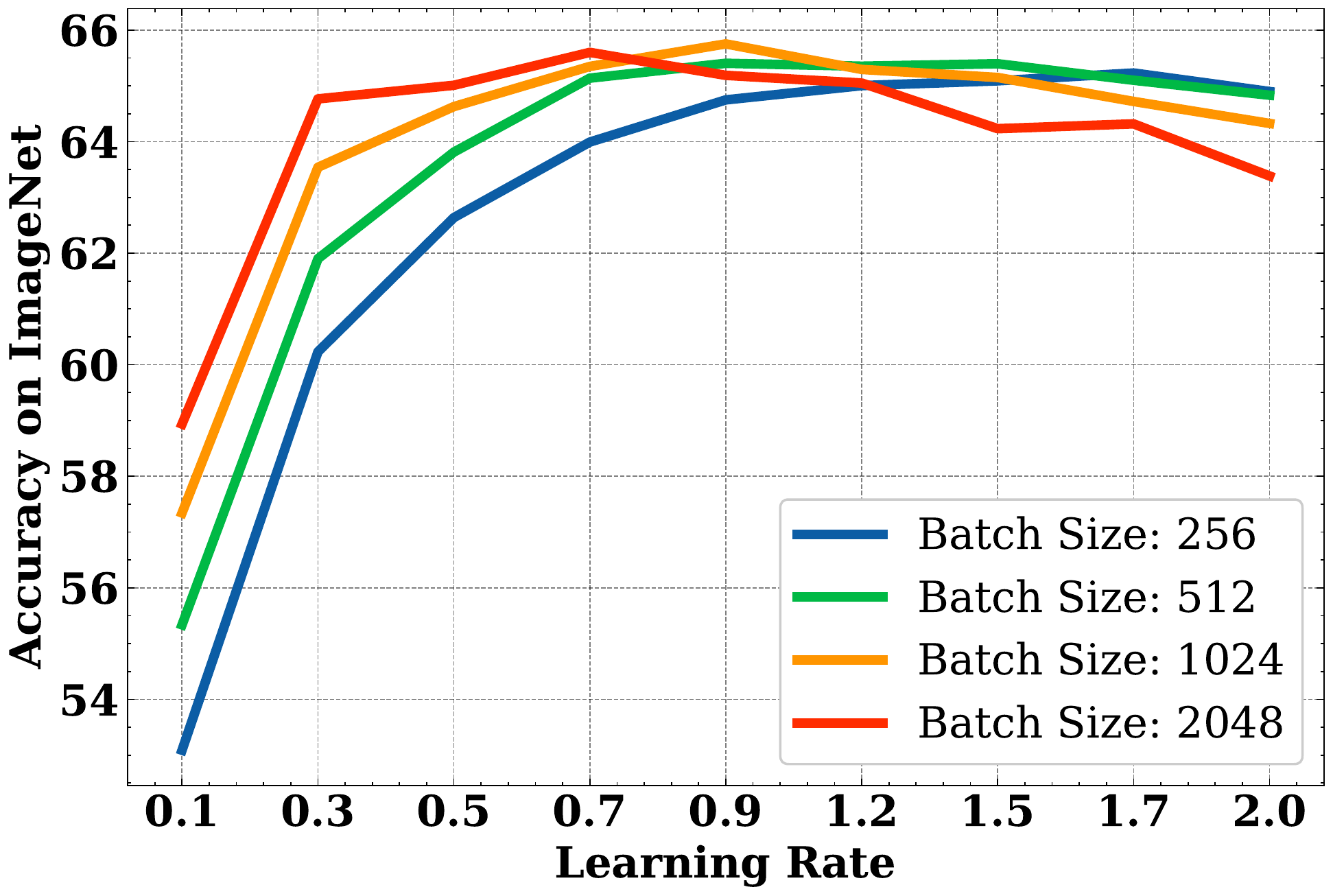}
    \vspace{-0.2cm}
    \caption{Validation Accuracy on ImageNet with respect to the learning rate (with LARS\cite{lars} as optimizer) for SimCLR. When having a learning rate of 0.3, the gap between a large batch size can be as high as 4\%. However, when carefully running a grid search, we can see that the optimal learning rate might not be the same depending of the batch size.}
    \label{fig:simclr_lr}
\end{figure}

\begin{figure}
    \centering
    \includegraphics[scale=0.4]{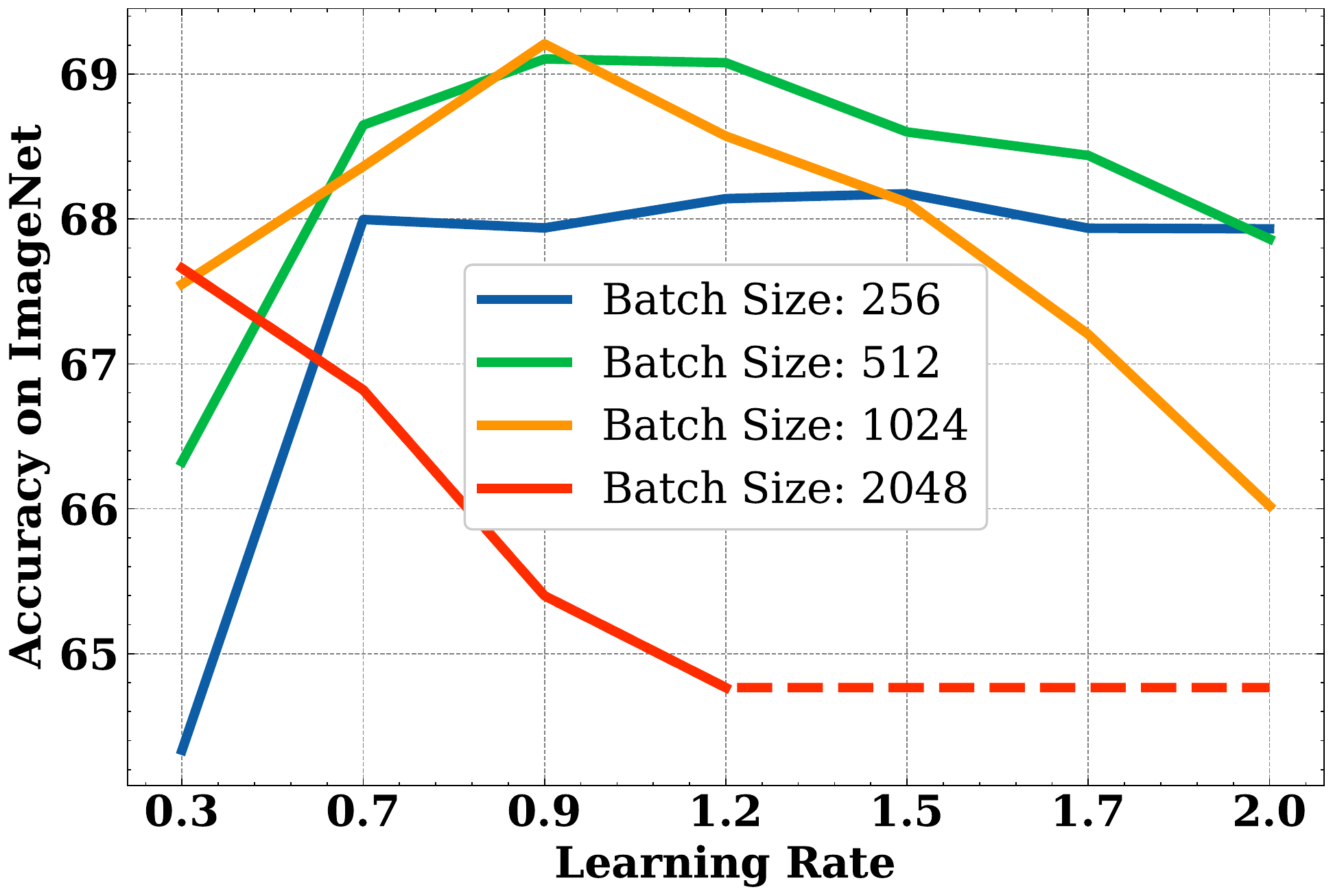}
    \vspace{-0.2cm}
    \caption{Validation accuracy on ImageNet with respect to the learning rate (with LARS\cite{lars} as optimizer) for Barlow Twins. Like SimCLR, the optimal learning rate can be radically different depending the batch size one had decided to use to train the model. So, it's really important to perform a grid search over the learning rate when changing the batch size. The dashed line corresponds to the situations when training Barlow Twins resulted in nan.}
    \label{fig:barlow_lr}
\end{figure}

\paragraph{The impact of the learning rate}  Another important parameter that interacts with the batch size is the learning rate. In \cref{fig:simclr_lr}, we show the validation accuracy on ImageNet of a SimCLR trained with different batch sizes and learning rates using the LARS optimizer\cite{lars}. We observe a significant impact on the gap in accuracy between larger and smaller batch size. When taking a single point as learning rate, for example $0.1$ we observe a $4\%$ accuracy gap between a batch size of 2048 and 256. However, if we carefully tune the learning rate, the gap becomes much smaller. We present a similar experiment with Barlow Twins in \cref{fig:barlow_lr}. There we also observe the high sensitivity to the learning rate depending of the batch size. For Barlow Twins, we even see the training become very unstable when using high learning rate with large batch size. 

\paragraph{The impact of Guillotine regularization} We further explore the batch dependency with respect to the number of layers in the projector. Since \cite{guillotine} demonstrated that the main role of the projector in SSL methods is to absorb the bias of an ill-defined training pre-text task, we hypothesize that using a deeper projector might help in reducing the gap in performances between different mini batch sizes. In Table \ref{tab:simclr_projector_impact}, we show the results of training SimCLR with a different number of layers in the projector. Adding layers in the projector indeed helps bridge the gap in performances between a large and a smaller mini batch size.

\begin{table}[t!]
    \centering
    \begin{tabular}{c|c|c|c|c|c}
          \hline
         BS/Layers & 1 & 2 & 3 & 4 & 5 \\
         \hline
         128 & 57.8 & 66.8 & 66.8 & 66.8 & 66.8  \\
          \hline
         256 & 59.3 & 66.4 & 68.1 & 68.4 & 68.4  \\
          \hline
         512 & 60.2 & 67.9 & 69.6 & 69.5 & 69.5  \\
          \hline
         1024 & 61.3 & 69.3 & 70.3 & 70.3 & 70.5  \\
          \hline
         2048 & 62.0 & 69.7 & 70.7 & 70.5 & 70.5  \\
    \end{tabular}
    \caption{Effect of the number of layers in the projector on the ImageNet validation accuracy depending on different batch sizes with SimCLR. For this setting, we use the same learning rate for all models. We use a non linear multilayer perceptron (two layers) as representation probing and present the best validation accuracy for each model. In this table, we observe directly that one can gain several accuracy percentage point in adding layers in the projector even for small batch sizes. The main takeaway is that \textbf{when comparing SSL methods, one should use the same number of layers in the projectors.}}
    \label{tab:simclr_projector_impact}
\end{table}

\subsection{The Impact of Data-Augmentation}

\begin{figure}
    \centering
    \includegraphics[scale=0.4]{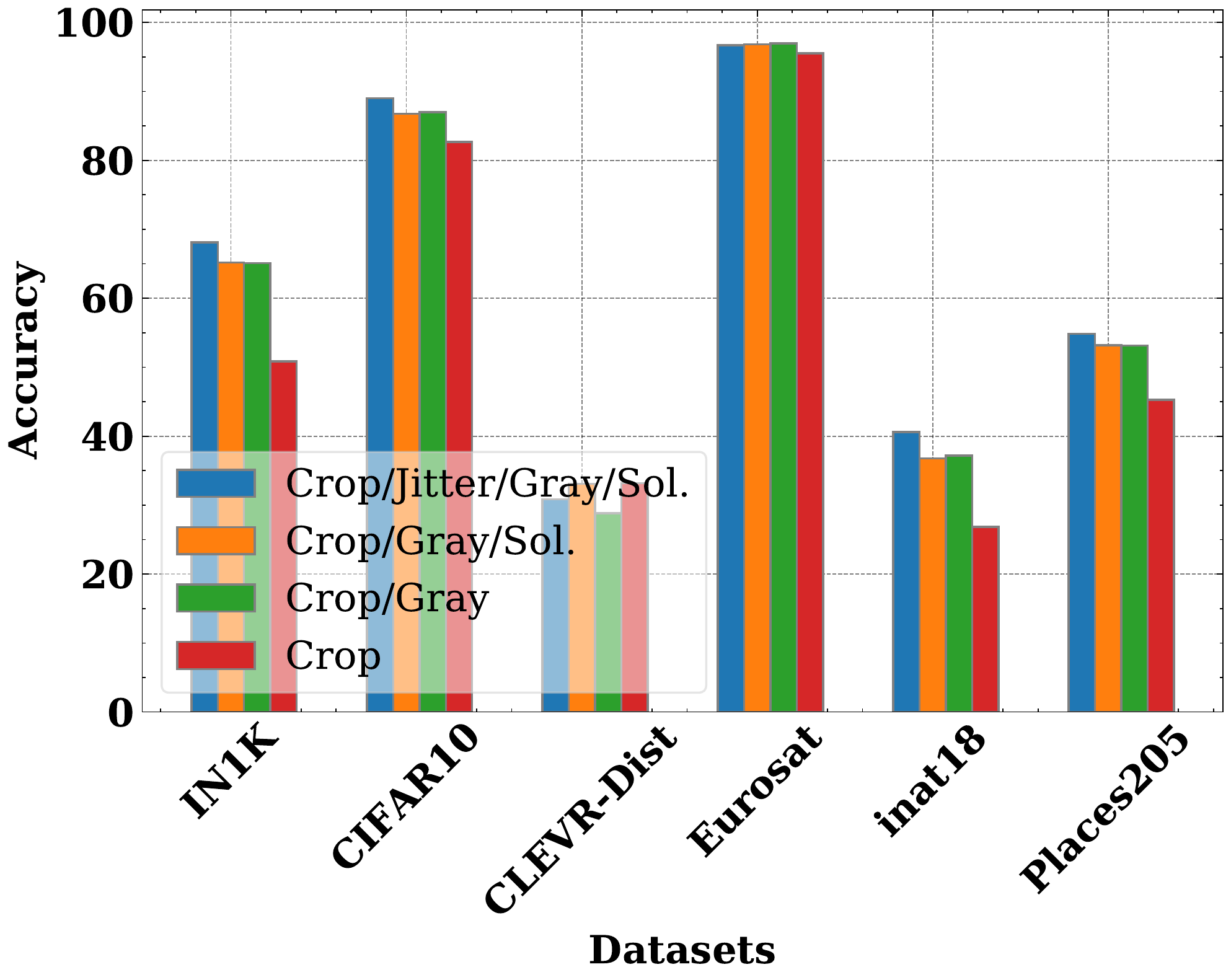}
    \vspace{-0.2cm}
    \caption{Accuracy for various downstream tasks for a model trained with SimCLR with different data augmentations. \textit{So} corresponds to a Solarization transformation with is applied with a 20\% probability, \textit{Gray} corresponds to a Grayscale operation that is also applied with a 20\% probability, \textit{B.} corresponds to a gaussian blur that is applied 100\% of the time and \textit{Jitter} is the ColorJitter operation, often used in SSL with a probability of 80\%. For most of the downstream tasks, there is an important gain in accuracy using all the sets of augmentations available. Surprisingly, the performances in only using cropping and a simple grayscale. transformation is very competitive. }
    \label{fig:SimCLR_change_DA_augs}
\end{figure}

Another important point that is often discussed about JE-SSL methods is their need for very strong data augmentations like Color Jitter \cite{chen2020simclr, zbontar2021barlow}. However, ColorJitter comes with an important computational cost (see \cref{table:torch_time}). To determine how much these augmentations are important with respect to classification performances, we follow a similar experimental protocol as in the previous section, except that we study the impact of specific data augmentations on different downstream tasks. In  \cref{fig:SimCLR_change_DA_augs}, we perform this experiment on SimCLR. For some of the classification downstream tasks, there is a important benefit in using all the data augmentations. But for others it is not as obvious. Surprisingly, we found that a simple cropping with a simple grayscaling on one branch gives very good performances with the benefit of being much computationally much cheaper than the ColorJitter operation (\cref{table:torch_time}).

\begin{figure}
    \centering
    \includegraphics[scale=0.4]{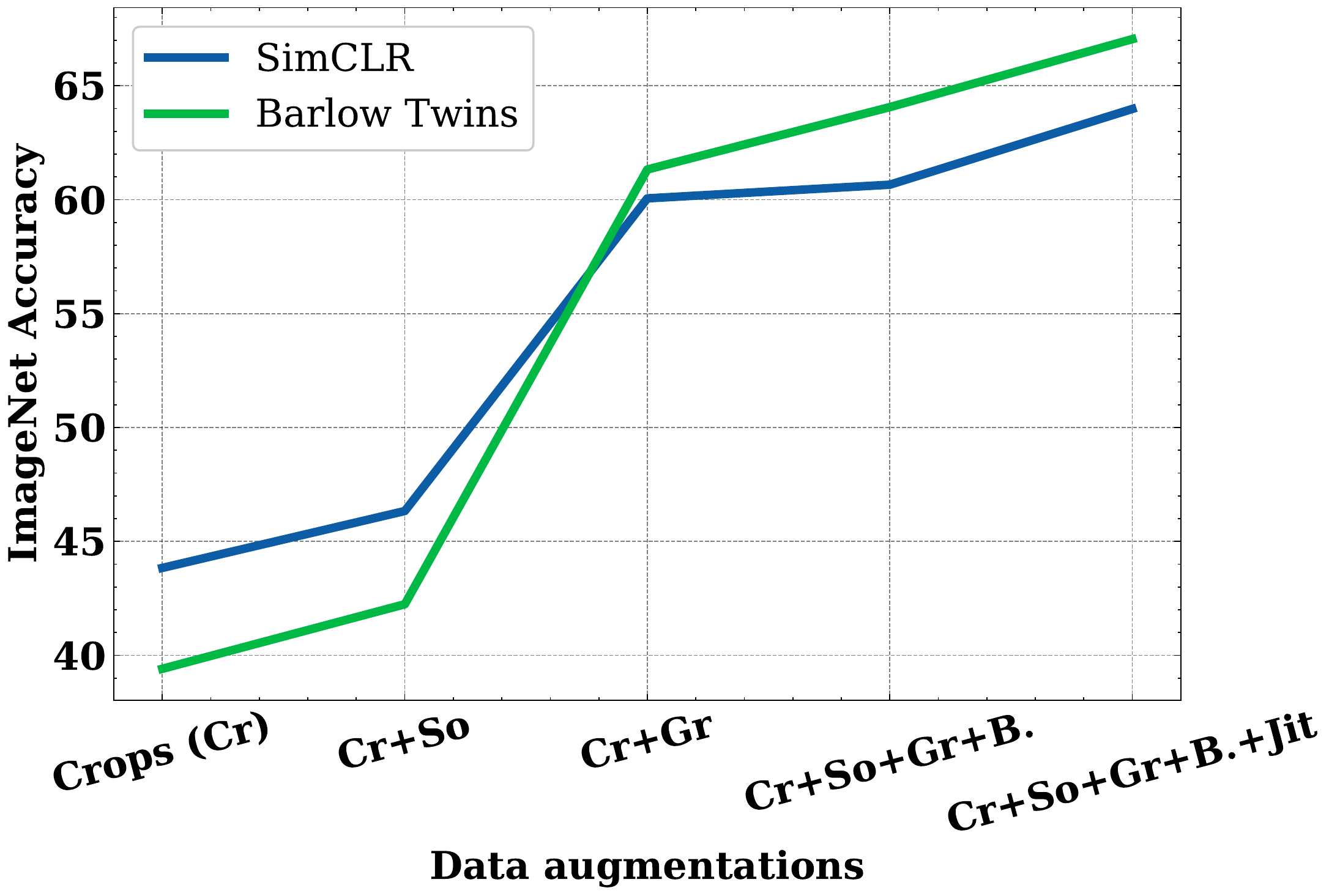}
    \vspace{-0.2cm}
    \caption{Detailed impact of the data augmentations used during SimCLR and Barlow Twins training on the ImageNet validation accuracy. As in \cref{fig:SimCLR_change_DA_augs}, \textit{So} corresponds to a Solarization transformation applied with a 20\% probability, \textit{Gray} corresponds to a Grayscale operation that is also applied with a 20\% probability, \textit{B.} corresponds to a Gaussian blur applied 100\%of the time and \textit{Jit} is the ColorJitter operation with $80\%$ probability. In this Figure, we can clearly see that the addition of grayscaling have the most significant impact on the ImageNet accuracy. }
    \label{fig:simclr_barlow_augs}
\end{figure}

This observation also holds for other SSL-JE methods. Figure \ref{fig:simclr_barlow_augs}, shows the ImageNet validation accuracy obtained by SimCLR and Barlow Twins using different data augmentations. We use similar data augmentations as the ones presented in \cref{fig:SimCLR_change_DA_augs}. Similarly, we observe that the grayscaling operation appears key for both SimCLR and Barlow Twins. Consequently, it seems unfair to discard JE-SSL methods due to their supposed need for strong and rich data augmentation, as merely cropping and grayscaling is able to yield good performances. 

\subsection{The impact of the evaluation protocol}
\begin{figure}
    \centering
    \includegraphics[scale=0.4]{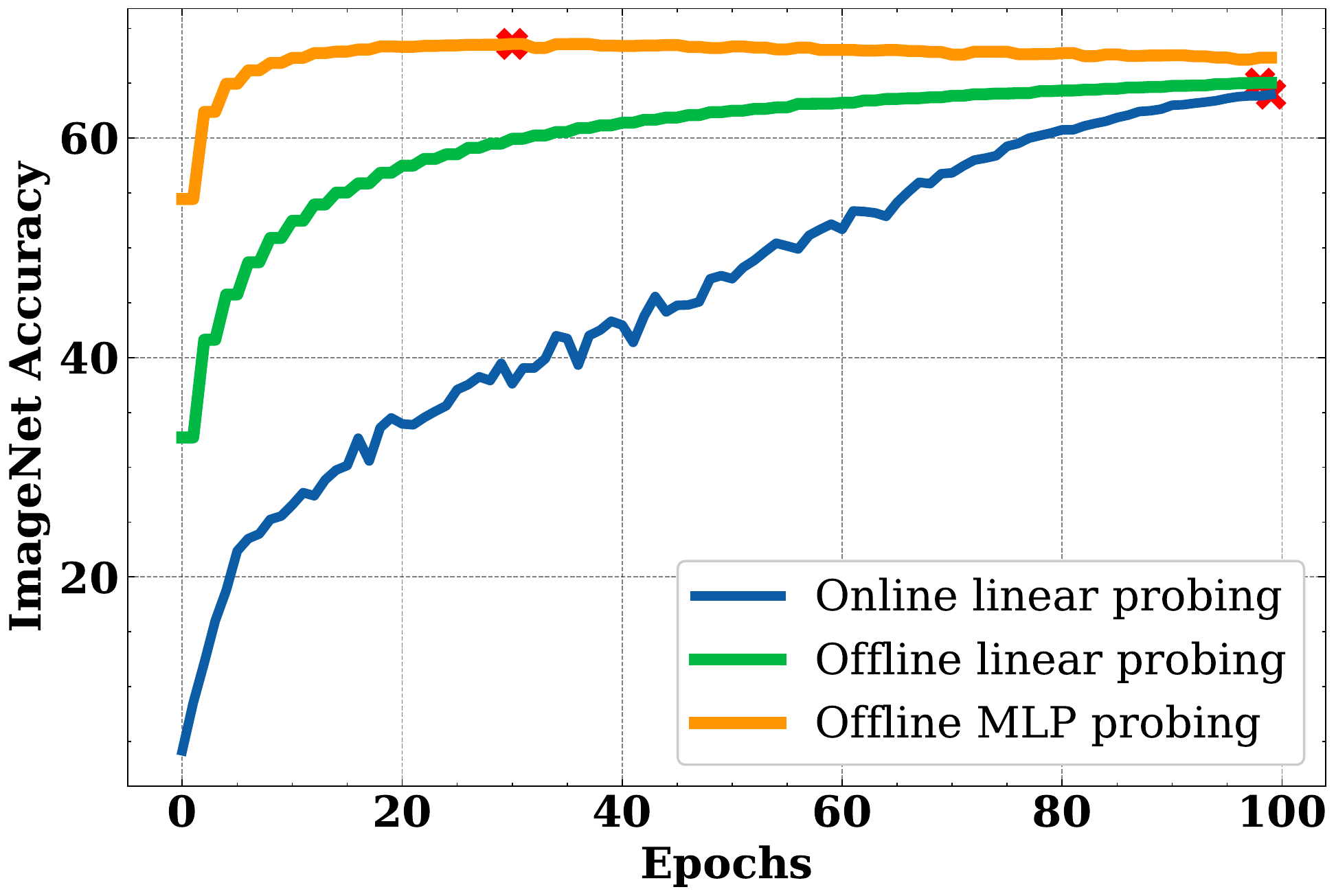}
    \caption{Depiction of the classifier probe trained to predict the Imagenet-1k labels from the output of the backbone during training ({\bf online}) and post-training ({\bf offline}) using a linear or MLP classifiers. The cross in red corresponds to the best accuracy. In the offline setting no data-augmentation is employed. We observe clearly that (i) when employing an MLP only a few epochs are needed and regularization or early-stopping should be employed, however, in the popular linear case, we clearly see that there is limited differences between the online and offline performances, and that over-fitting never occurs during either of the training cases. 
    }
    \label{fig:linear_vs_mlp}
\end{figure}

Another important dimension in Self-Supervised learning is the protocol for evaluating the learned representation, which is typically limited to linear probing and/or fine-tuning, occasionally complemented by qualitative visualizations such as RCDM \cite{RCDM}. A more thorough evaluation of the quality of the learned representation and its suitability for downstream tasks should also consider non linear probing.
In \cref{fig:linear_vs_mlp}, we compare the performances with linear probing in an online scenario (meaning training the linear probe at the same time as training the SSL model with a cut gradient on the classifier input) and an offline setting (when training the linear probe occur only after training) as well as an offline non linear probing (a two layers MLP of size 2048-2048-1000). We first observe that the performances between online and offline linear probing are well correlated. When using a non linear probing, we observe a significant boost in accuracy, however the optimal validation score is not necessarily the one at the last epoch. This is not surprising  since it is much easier for a non linear probe than a linear probe to overfit on its training set.

\subsection{Per-Instance Positive and Negative Sample Generation}

Most JE-SSL methods, and in particular contrastive ones such as SimCLR, share a need to have positive (semantically similar) and negative (dissimilar) inputs. 
The latter prevent the representation from collapsing, and their quality is thus a predominant concern in JE-SSL. This e.g. motivated the idea of hard-negative sampling. 
In this section, we propose to debunk the idea that it is necessary for the model to work hard (i.e. consider many other examples) to get good negatives. 
We showcase the ability to generate useful negative samples for each image, based on simple transformations of only that \emph{same} image.

\begin{figure}
    \centering
    \includegraphics[scale=0.4]{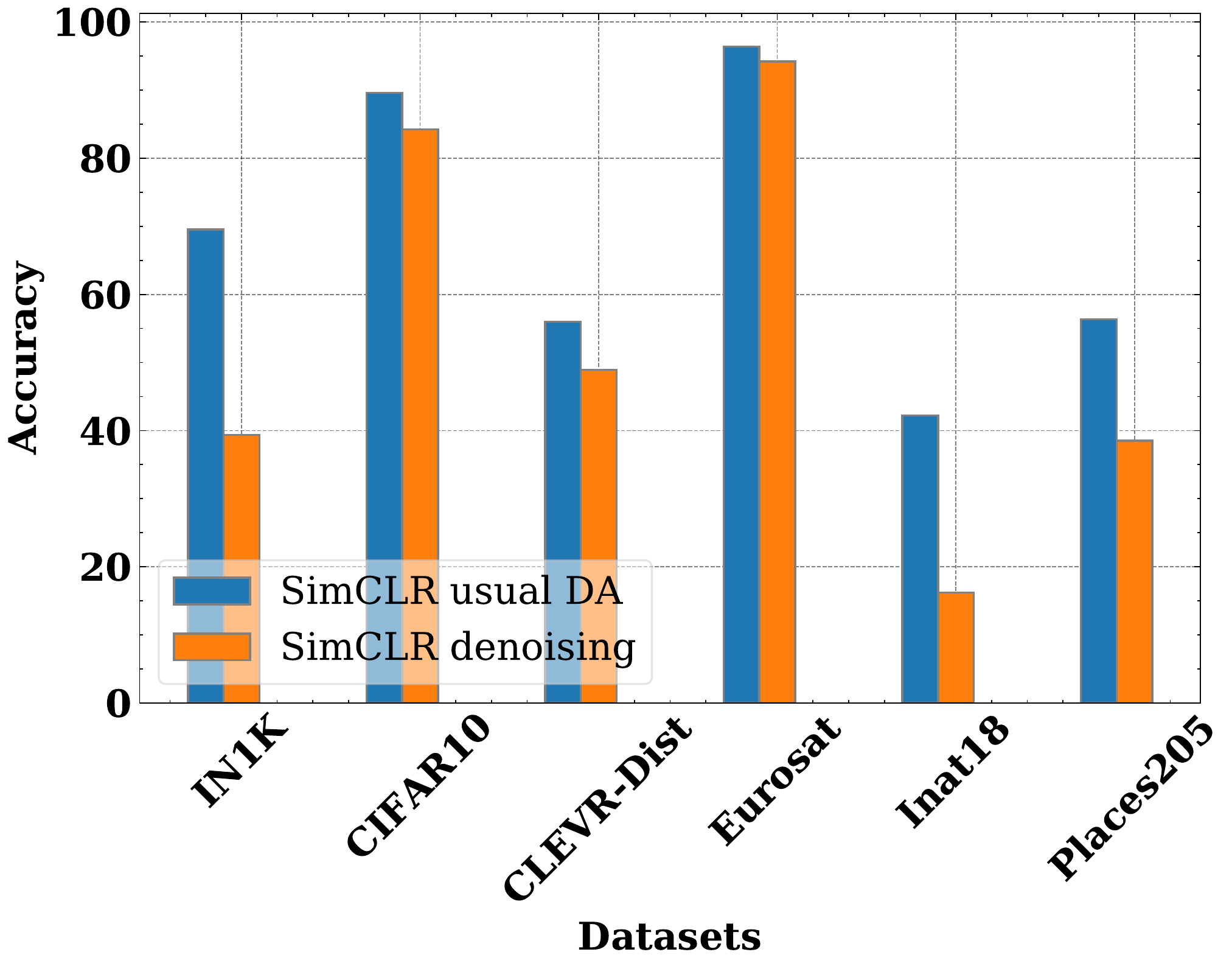}
    \caption{Accuracy on various downstream tasks using an instance based SimCLR (SimCLR denoising in the figure) and a SimCLR using the usual data augmentations. Instead of using all the examples in the mini batch as negatives examples, we decided to use only a small crop of the positive pair as negatives examples. We also replaced all the SSL data augmentations by only a simple Gaussian noise, thus instead of learning to be invariant to cropping or colorjitter, this instance SimCLR learn only to denoise a specific patch of the image. Doing so allow the network to avoid collapse while having competitive accuracy on downstream tasks like CIFAR10 or the Eurosat dataset. The drop in accuracy on ImageNet is expected since we reduce a lot the amount of inductive bias since we only used Gaussian noise as data augmentation without cropping. }
    \label{fig:instance_simclr}
\end{figure}

Specifically we attempt to train SimCLR only using a single random patch as negative examples while using only Gaussian noise as data augmentation to define the positive pair. This set-up deliberately defies most current guidelines, which favor aggressive data-augmentations and hard-negative sampling. However, despite that fact that we will leverage the same instance to define the positive and negative pair, it shouldn't be  too surprising that different crops can produce relatively different images. In \cref{fig:instance_simclr} we show the accuracy across diverse downstream tasks obtained with such an instance based SimCLR model. Surprisingly this somewhat extreme approach  can nevertheless learn good representations useful for several downstream tasks. We get high accuracy on CIFAR10 or Eurosat, but only $40\%$ on ImageNet. The performance drop on ImageNet is not surprising given we reduced a lot the amount of inductive bias in the augmentations. However, the high performances on transfer on CIFAR10 is still surprising. We hope that our work will motivate further the exploration of such extreme scenario.

\vspace{-0.2cm}
\section{FFCV-SSL: A Fast Data Loading Library Tuned to Improve JE-SSL Training Time}
\label{sec:FFCV_main}

Most implementations of Self-Supervised learning use Pytorch \cite{pytorch} along its vision library Torchvision. Since SSL methods rely on an important set of rich data augmentations, we hypothesized that the data loading process could become an important  bottleneck when training SSL models. To verify our hypothesis, we replace Torchvision by the data loading FFCV \cite{leclerc2022ffcv}
 library. More precisely we created a fork of this library that we called FFCV-SSL, to include most of the data augmentations that are currently used in Self-Supervised learning. We also added the ability to manage multiple pipelines in parallel (which is needed in SSL-JE since we have at least two different views of a given image). 

We started by investigating the time consumed to only fetch the data. In \cref{table:torch_time}, we show the time to pass through the entire ImageNet dataset (when using no neural networks) with the torchvision and FFCV-SSL dataloader for various data augmentations. We can see that adding  augmentations significantly increases the time for a full epoch. The addition of the traditional SSL augmentations leads to almost a two times slow down, only for data processing.

\begin{table}[t!]
\setlength\tabcolsep{0.5em}
\begin{center}
\begin{tabular}{c|c|c|c|c|c} 
 \multicolumn{1}{c}{}& Crops & +Blur & +Gray. & +Sol. & +Jitter \\ 
 \hline
 Torchvision & 7:00 & 9:25 & 9:26 & 9:30 & 13:20 \\ 
 \hline
 FFCV-SSL & 1:30 & 1:36 & 1:58 & 2:07 & 7:00 \\ 
 \hline
\end{tabular}
\end{center}
\caption{Time (minutes: seconds) for one complete epoch over the torchvision data loader for various data augmentations. We observe that the Blur and ColorJitter operation add a considerable time in the training.}
\label{table:torch_time}
\end{table}

However, measuring only the data loading time can be misleading, since the forward and backward pass of the model in a training loop will take additional times that can play with the caching process. When measuring the training time for a single epoch with torchvision, using the full set of data augmentations, we found that it takes on average \textbf{1141s} for one epoch while FFCV take around \textbf{428s}. This shows that merely switching the data loading library can give us almost a three times speed up, for a single epoch. 

In Figure \ref{fig:ffcv_vs_torchvision}, we plot the accuracy on the ImageNet dataset with respect to the training time for models trained with FFCV-SSL and torchvision. Like the original FFCV, FFCV-SSL has the ability to easily switch the resolution of the data during training, which can also significantly improve the training time. Using FFCV-SSL, we can train a SimCLR in less than 8 hours using 8 GPUs A100. When using a single GPU (one A100), the training time take 35 hours which is close to the training time of SimCLR using torchvision when trained on 8 GPUs. Having a way to faster train SSL-JE models on 8 or even only 1 GPU will help democratize this research area. This will also enable more thorough hyper-parameter search, which should avoid the pitfall of drawing flimsy scientific conclusion based on too few data points. 
In addition, the codebase we developed allows to easily perform training and evaluation of different SSL models. To straightforwardly guarantee using the same experimental setup, -- which is important when comparing different methods -- we wrote a simple file that supports many existing SSL losses. In contrast with complex libraries such as VISSL \cite{goyal2021vissl}, our main code is self-contained in a single file. This makes it easier to hack and to tune every-hyper parameters that researchers might need.

One limitation of our work is that, as shown in \cref{table:torch_time}, the data augmentation still take a lot of time, especially Grayscale and ColorJitter. One should probably be able to get additional speed up by using further optimized data transformations.

\subsection{Enabling single GPU training with FFCV-SSL}
\label{sec:single_GPU}

As demonstrated in the previous section, the necessity of strong augmentation and large batch size in SimCLR is not as obvious as  presented in the current literature. Knowing this, we attempted to train several SimCLR models using a \emph{single} GPU.

\begin{figure}
    \centering
    \includegraphics[scale=0.4]{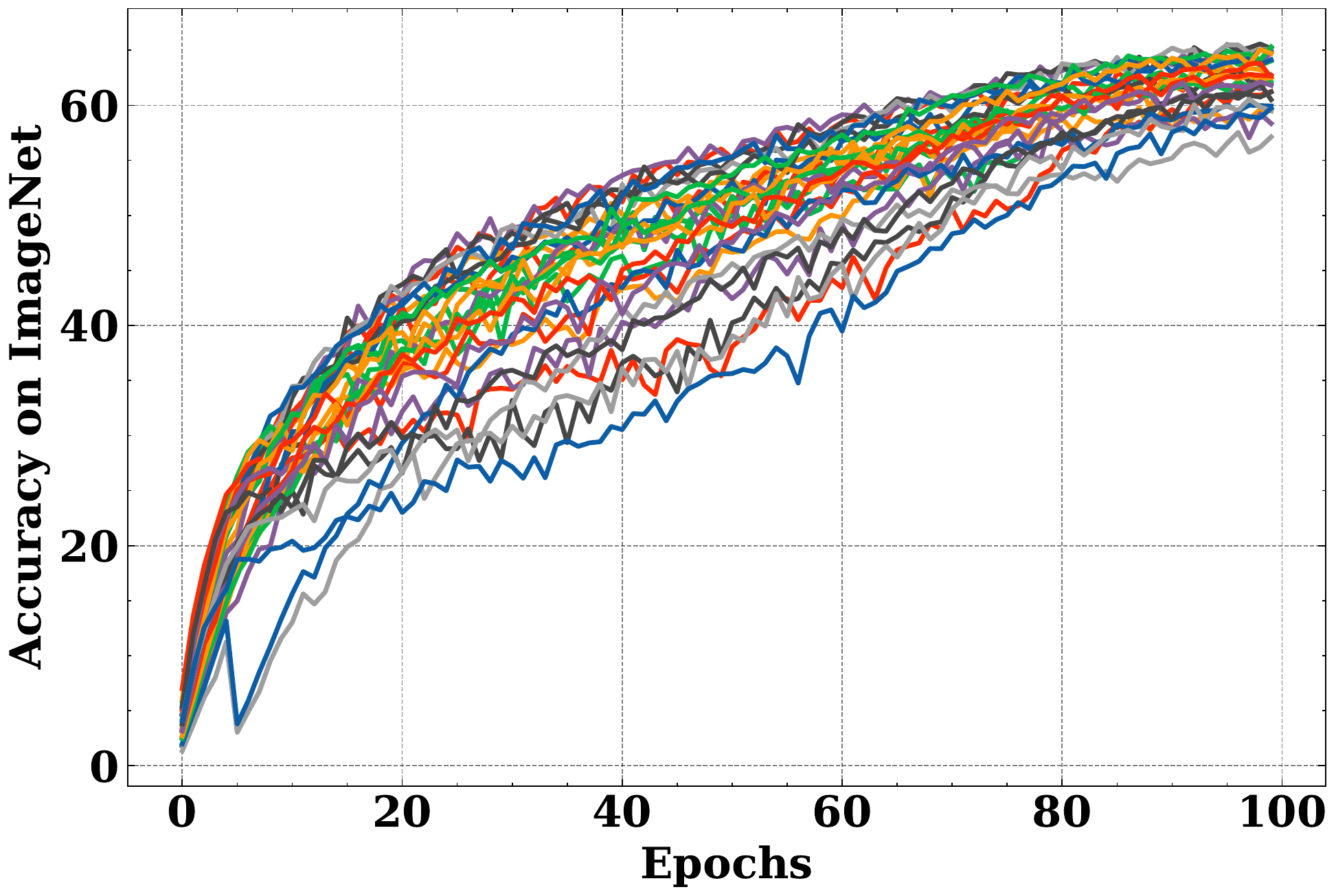}
    \caption{Imagenet validation accuracy on a wide cross validation performed on a single gpu with SimCLR. For this experiment, we performed a grid search on the temperature with the values $0.10, 0.15, 0.25, 0.5$ and the learning rate with the following values: $0.3, 0.5, 0.7, 1.0, 1.2, 1.5, 2.0, 2.5, 3.0$. We found that the best hyper-parameters for a Single GPU training, using a batch size of 256, are a temperature of $0.15$ and a learning rate of $1.0$, this lead to a $65.58$ accuracy in online linear probing and $68.5\%$ in non linear probing. } 
    \label{fig:CV_SimcLR_1gpu}
\end{figure}

Equipped with the novel insights from \cref{sec:debunking}, we are now able to provide a minimal example of JE-SSL that enables single GPU training on Imagenet-1k. In \cref{fig:CV_SimcLR_1gpu}, we performed an extensive grid search on the temperature and learning rate parameters. We show that the best hyper-parameters for a single GPU training, using a batch size of 256, are a temperature of $0.15$ and a learning rate of $1.0$, this lead to a $65.58$ accuracy in online linear probing and $68.5\%$ in non linear probing. Even if the accuracies we achieved are much higher than what has been showed in the literature for small batch size, there still remains a small unexplained gap in performances on ImageNet. 

\section{Recalibrated Observations for Self-supervised learning research}

Many misleading ideas about the importance of batch size and data augmentations were widely shared and spread among JE-SSL studies without challenging them due to computational costs. Since this cost can be significantly alleviated by the use of a better library for loading the data, we are now able to push back on some of those ideas and to provide a clearer  recalibrated view of our findings:
\begin{itemize}
    \item When training SSL-JE models, adverse impact of the batch size on downstream performances can be largely countered by adapting the learning rate accordingly.
    \item The need for specific strong data augmentations is not clear. It is possible to reach $63.5\%$ top-1 on Imagenet-1k using only grayscale and cropping which is also a computationally cheaper set of augmentation to employ.
    \item Imagenet-1k top-1 metric does not contain the full picture when measuring the quality of a learned representation, we believe that the considered set of OOD tasks in \cref{fig:simclr_lr} is more representative. 
    \item The projector architecture also plays a crucial role for final performance, and one should thus strive to compare JE-SSL methods with the same projector network instead of the official one that was found independently by each of the methods.
\end{itemize}

\vspace{-0.2cm}
\section{Conclusion}

We provided a thorough series of experiments aiming at debunking some popular but misconceived ideas around joint-embedding SSL. Because most studies capitalize on prior existing ones to limit the computation burden of training JE-SSL models, we found that many pre-conceived ideas had remained unchallenged for years, thus impairing the development of novel methods. For example we deliver key findings regarding the requirement of rich data-augmentations or the impact of mini-batch size. In additional to these findings, we provide key strategies to speed up training and evaluation e.g. using an online classifier probe, or replacing the costly color jittering augmentation with the much cheaper grayscale one. We developed a dedicated PyTorch library FFCV-SSL, based on FFCV \cite{leclerc2022ffcv}, that further enables rapid training of JE-SSL models e.g. producing on a single GPU faster training than an 8-GPU set-up with the usual torchvision pipeline. We hope that the collection of rectified findings and software that this study produced will enable a much broader deployment and seamless research development of JE-SSL.

{\small
\bibliographystyle{ieee_fullname}
\bibliography{main}
}

\newpage
\appendix
\label{sec:FFCV_app}

\section{FFCV-SSL: A library for fast SSL training}

In this paper, we introduced FFCV-SSL, a fork of the FFCV library \cite{leclerc2022ffcv} that we improved to make training of SSL models much faster. FFCV increases significantly the speed of data loading by converting any image dataset to a single file with fixed or variable resolution for each images. In addition, all data augmentations are compiled in advance with Numba. In our implementation, we added the following data augmentations: ColorJitter/Grayscale/Solarization and also the support for multiple branch of augmentation given a specific input.
The library is available at \url{https://github.com/facebookresearch/FFCV-SSL}. We added a code file in the supplementary material that show how to use FFCV-SSL to train several type of SSL methods. Using this single file one can train SimCLR/VICReg/BarlowTwins and Byol using a single or many gpus. The script support the use of SLURM with submitit.

\section{On the importance of increasing the learning rate when using small batch size}
\begin{figure}
    \centering
    \includegraphics[scale=0.4]{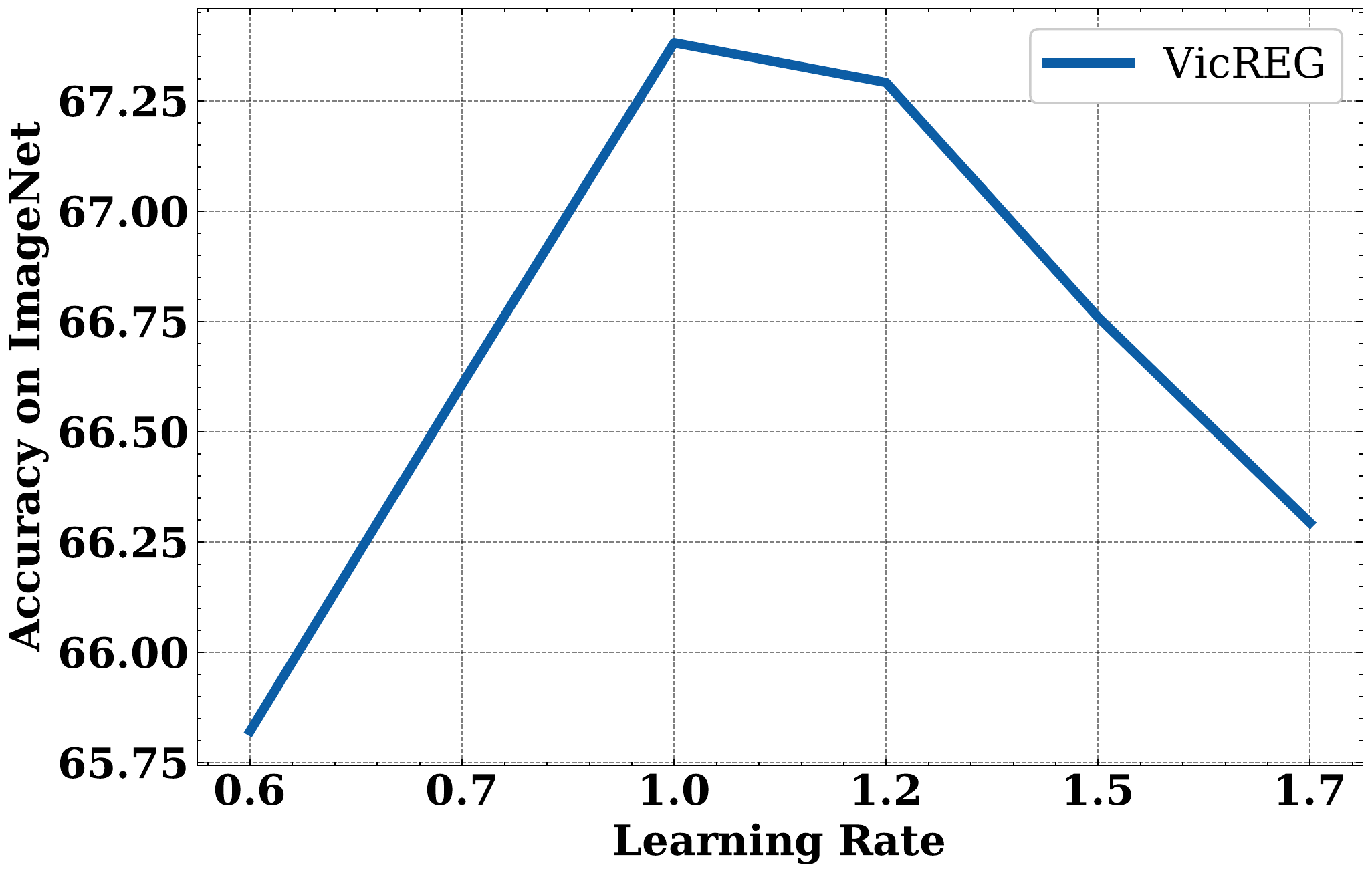}
    \caption{Imagenet validation accuracy with respect to the learning rate  on a single gpu (batch size 256) with VICReg.} 
    \label{fig:VicReg_lr}
\end{figure}

\begin{figure}
    \centering
    \includegraphics[scale=0.4]{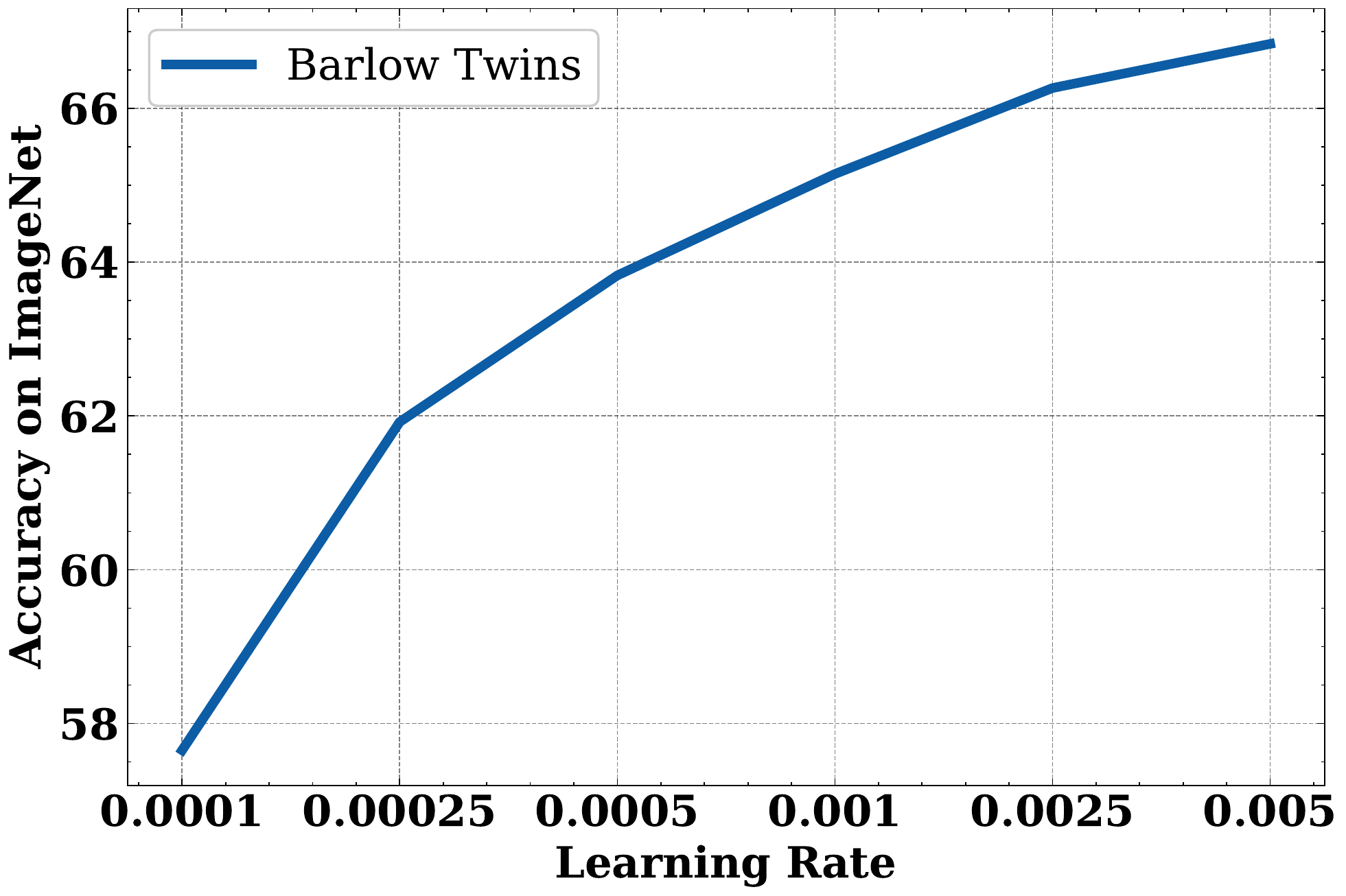}
    \caption{Imagenet validation accuracy with respect to the learning rate  on a single gpu (batch size 256) with Barlow Twins.} 
    \label{fig:Barlow_lr}
\end{figure}

\begin{figure}
    \centering
    \includegraphics[scale=0.4]{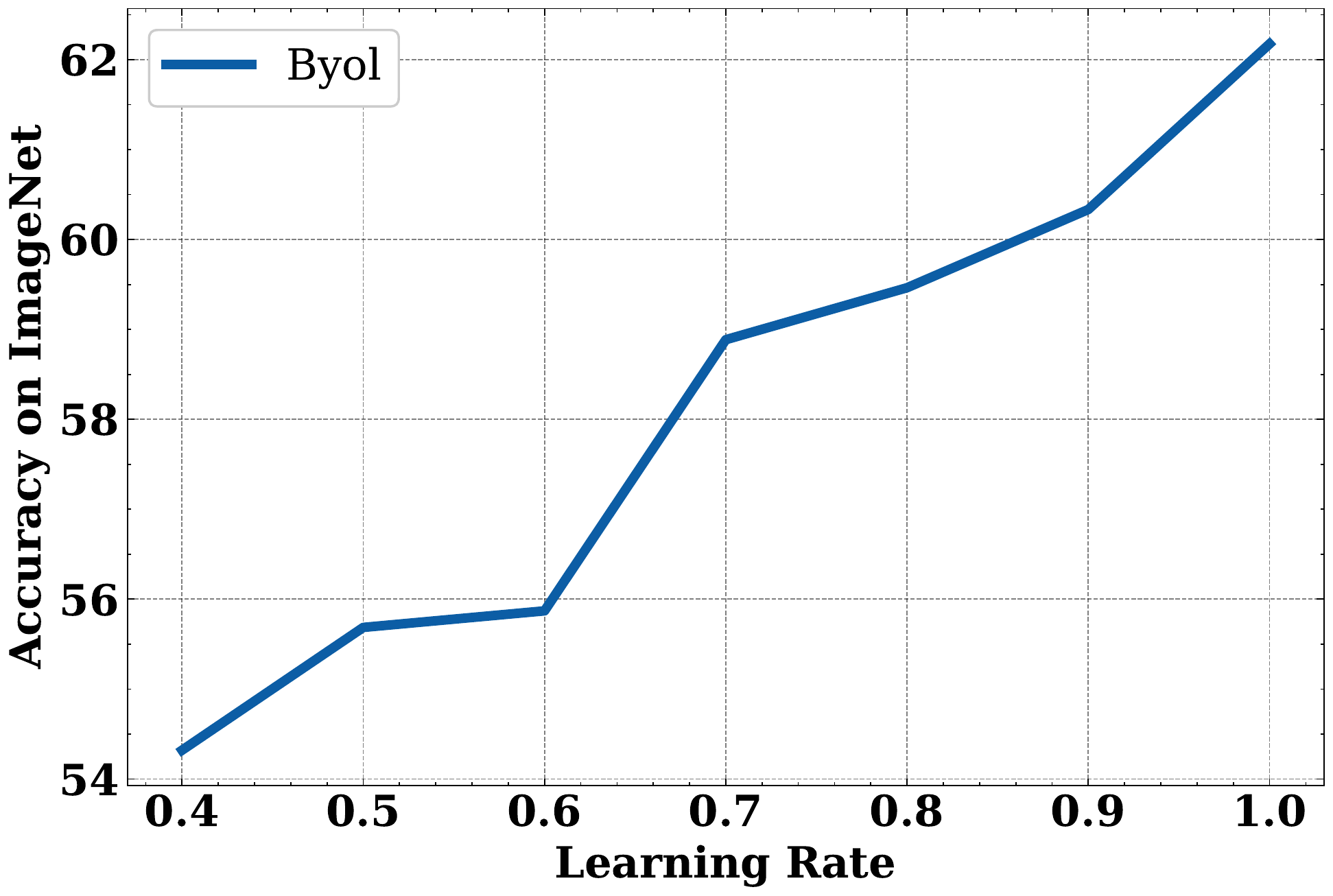}
    \caption{Imagenet validation accuracy with respect to the learning rate  on a single gpu (batch size 256) with BYOL.} 
    \label{fig:Byol_lr}
\end{figure}

In this section, we present more experiments concerning the impact of the learning rate when using small batch sizes. In \cref{fig:VicReg_lr}, we study the impact of the learning rate on VICReg when using a batch size of 256 on a single gpu. The optimizer used is LARS and we observe that one can get easily some percentage gains in accuracy by just tuning the learning rate. We observe the same behavior with Byol, using the same optimizer in \cref{fig:Byol_lr} and with Barlow Twins, using AdamW as optimizer in \cref{fig:Barlow_lr}.

\section{Additional Single GPU experiments}
\begin{figure}
    \centering
    \includegraphics[scale=0.4]{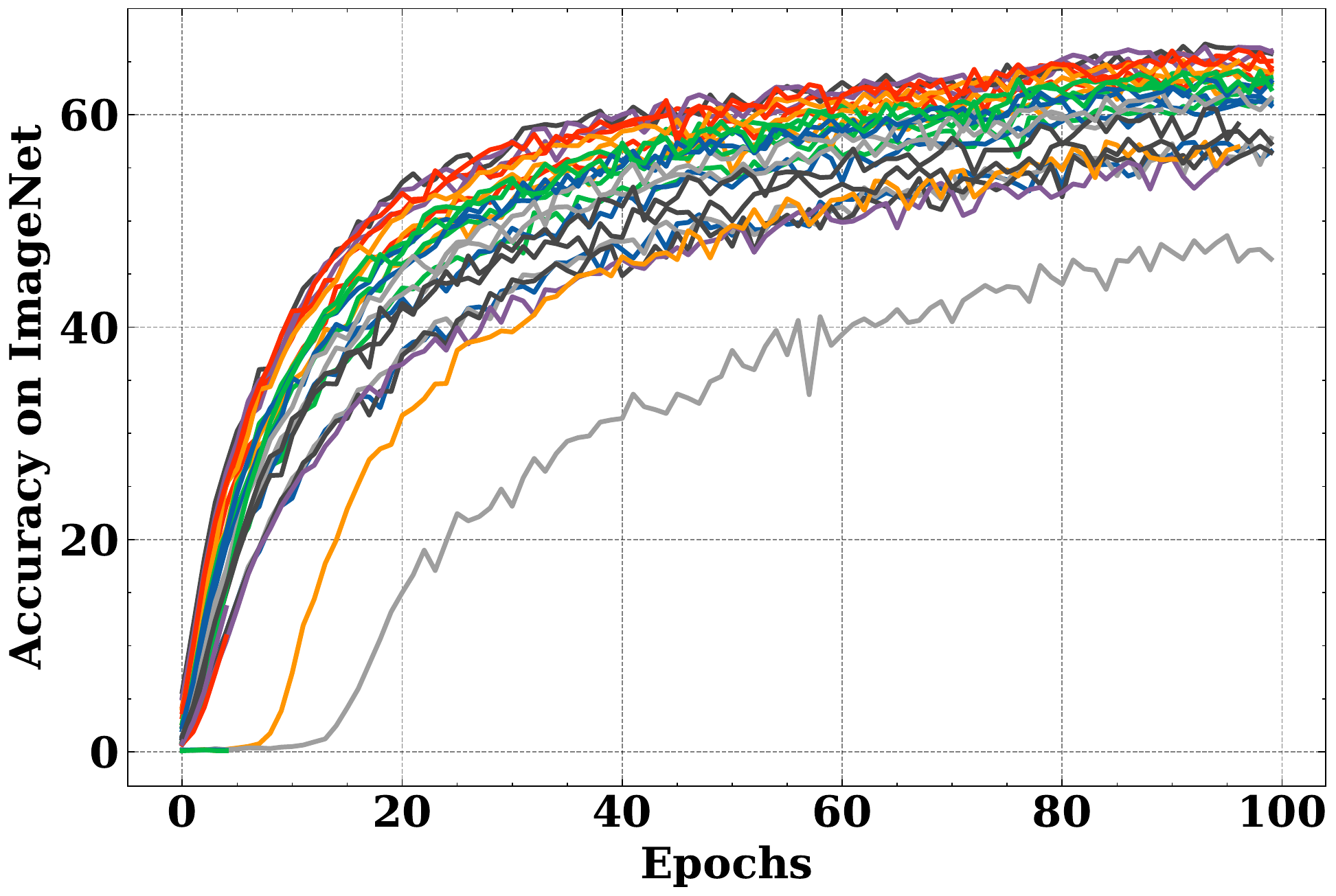}
    \caption{Imagenet validation accuracy on a wide cross validation performed on a single gpu with Barlow Twins. For this experiment, we performed a grid search on the Barlow Twins lambd hyper-parameter with the values $.0025, 0.0045, 0.0051, 0.0075, 0.01$ and the learning rate (AdamW with a weight decay of $4e-5$) with the following values: $0.0001, 0.00025, 0.0005, 0.001, 0.0025, 0.005$. We found that the best hyper-parameters for a single GPU training, using a batch size of 256, are a lambd value of $0.0025$ and a learning rate of $0.005$, this lead to a $66.8$ accuracy in online linear probing.} 
    \label{fig:CV_Barlow_1gpu}
\end{figure}

In this section, we present several grid search of hyper-parameters for different ssl methods when using a single gpu. In \cref{fig:CV_VicReg_1gpu}, we found that the optimal hyper-parameters for VICReg on a single gpu, based on this grid search are a similarity and std coefficient of $25$ and a learning rate of $1.0$ when using LARS as optimizer. With Barlows Twins in \cref{fig:CV_Barlow_1gpu}, we found that the optimal hyper-parameters on a single gpu is a lambd value of $0.0025$ and a learning rate of $0.005$ using AdamW. For Byol \cref{fig:Byol_lr}, we found the optimal hyper-parameter to be a momentum encoder value of $0.996$ and a learning rate of $1.0$.

\begin{figure}
    \centering
    \includegraphics[scale=0.4]{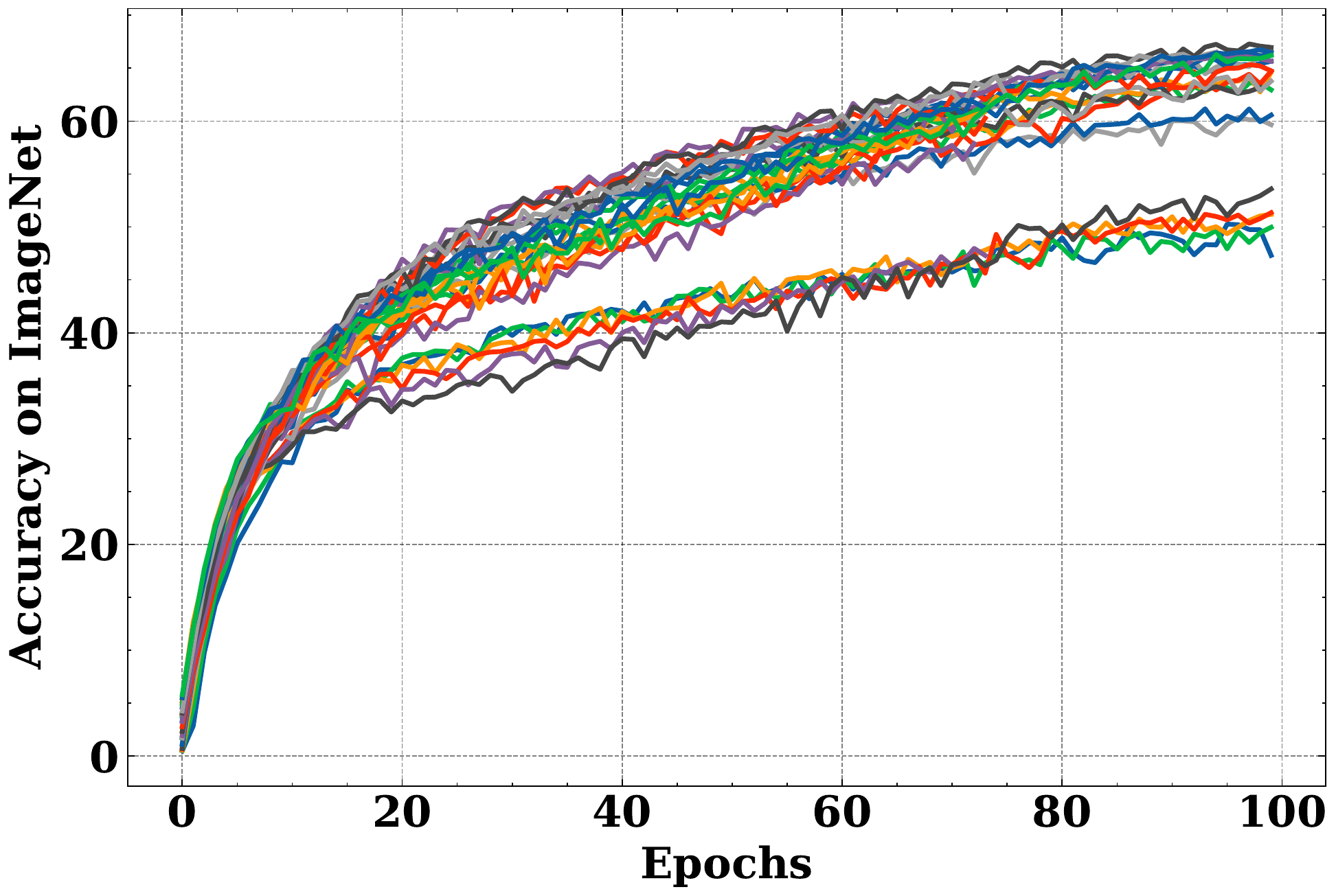}
    \caption{Imagenet validation accuracy on a wide cross validation performed on a single gpu with VICReg. For this experiment, we performed a grid search on the similarity and std coefficient with the values $1, 5, 10, 15, 25$ (We fixed the covariance coefficient to 1) and the learning rate (LARS with a weight decay of $1e-4$) with the following values: $0.6, 0.7, 1.0, 1.2, 1.5, 1.7$. We found that the best hyper-parameters for a single GPU training, using a batch size of 256, are a similarity and std coefficient of $25$ and a learning rate of $1.0$, this lead to a $67.4$ accuracy in online linear probing.} 
    \label{fig:CV_VicReg_1gpu}
\end{figure}

\begin{figure}
    \centering
    \includegraphics[scale=0.4]{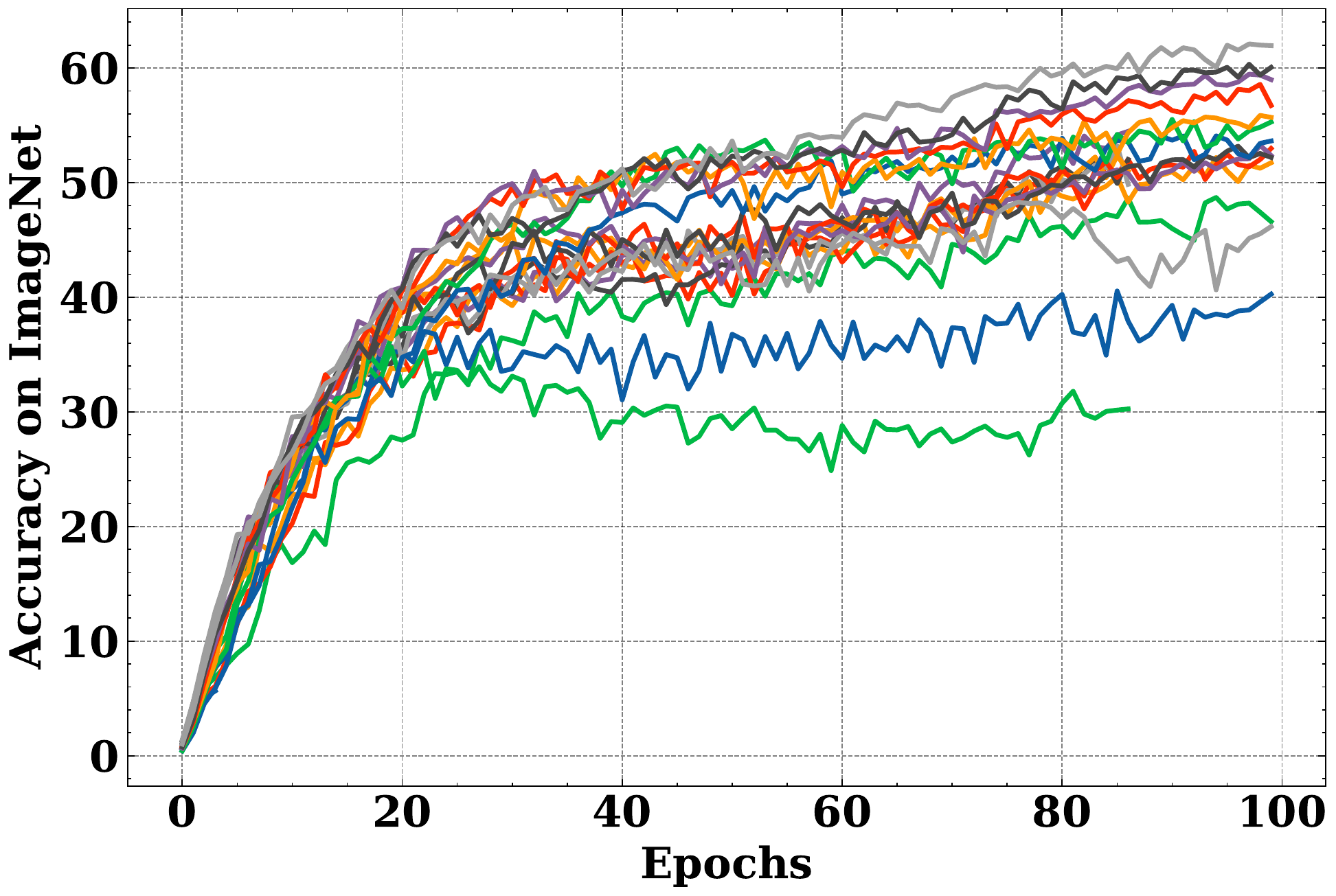}
    \caption{Imagenet validation accuracy on a wide cross validation performed on a single gpu with Byol. For this experiment, we performed a grid search on the momentum encoder hyper-parameter with the values $0.8, 0.9, 0.996$ and the learning rate (LARS with a weight decay of $1e-4$) with the following values: $0.4, 0.5, 0.6, 0.7, 0.8, 0.9, 1.0$. We found that the best hyper-parameters for a single GPU training, using a batch size of 256, are a momentum encoder value of $0.996$ and a learning rate of $1.0$, this lead to a $62.2$ accuracy in online linear probing. } 
    \label{fig:CV_VicReg_1gpu}
\end{figure}
\end{document}